\documentclass[conference]{IEEEtran}
\IEEEoverridecommandlockouts
\usepackage{cite}
\usepackage{amsmath,amssymb,amsfonts}
\usepackage{algorithmic}
\usepackage{graphicx}
\usepackage{textcomp}
\usepackage{xcolor}
\def\BibTeX{{\rm B\kern-.05em{\sc i\kern-.025em b}\kern-.08em
    T\kern-.1667em\lower.7ex\hbox{E}\kern-.125emX}}

\IEEEpubid{\copyright~2026 IEEE. Personal use is permitted. For any other purposes, obtain permission from the IEEE.}
    
\begin{document}

\title{ When Robots Rate Their Own Interactions:\\
Engagement Validity and the Strangeness Failure
}

\author{\IEEEauthorblockN{Victor Lockwood}
\IEEEauthorblockA{\textit{MS in AI Program} \\
\textit{Rochester Institute of Technology}\\
Rochester, New York, USA \\
vl3161@rit.edu}
\and
\IEEEauthorblockN{Hasan Mahmud}
\IEEEauthorblockA{\textit{School of Communication} \\
\textit{Rochester Institute of Technology}\\
Rochester, New York, USA \\
hxmdfp@rit.edu}
\and
\IEEEauthorblockN{Mohammad Javad Khojasteh}
\IEEEauthorblockA{\textit{Electrical and Microelectronic Engineering} \\
\textit{Rochester Institute of Technology}\\
Rochester, New York, USA \\
mjkeme@rit.edu}
\and
\IEEEauthorblockN{Prabu David}
\IEEEauthorblockA{\textit{School of Communication}\\
\textit{Rochester Institute of Technology}\\
Rochester, New York, USA \\
pxdpro@rit.edu}
\and
\IEEEauthorblockN{Jamison Heard}
\IEEEauthorblockA{\textit{Electrical and Microelectronic Engineering} \\
\textit{Rochester Institute of Technology}\\
Rochester, New York, USA\\
jrheee@rit.edu}
}

\maketitle

% 6-8 pages including references
% 

\begin{abstract}
Human-robot interaction (HRI) evaluation relies almost exclusively on human-completed questionnaires, leaving the robot's perspective unexamined. We propose an \textit{inverted evaluation}, in which LLM-powered robots complete the same standardized instruments from their own perspective, and test whether these ratings agree with human ground truth. In Study~1, five LLMs completed HRI-CUES, Godspeed, and RoSAS questionnaires for 25~interactions ($N = 1{,}522$ evaluations) from the HRI-CUES dataset. LLMs achieved moderate-to-strong agreement on engagement dimensions (satisfaction $r$ up to $.65$ and enjoyment $r$ up to $.72$) with excellent test-retest reliability (ICC $\geq .82$), but \textit{systematically inverted} the comfort/strangeness dimension ($r = -.44$ to $-.67$, all $p < .05$), conflating engagement with comfort. In Study~2, a Nao robot running Claude~Sonnet~4.5 replicated these patterns in live interactions ($N = 4$), including real-time turn-by-turn assessment. The strangeness failure persisted across five models, synthetic controls, and embodied deployment for two participants. We argue that current LLM-based robots lack access to the internal affective states needed to assess constructs like strangeness, and that inverted evaluation requires supplementary modalities (e.g., physiological signals, gaze, proxemics) to move beyond behavioral proxies. These findings establish boundary conditions for using LLMs as interaction evaluators in HRI.
\end{abstract}

\begin{IEEEkeywords}
human-robot interaction, large language models, interaction
evaluation, social robots, questionnaire methodology
\end{IEEEkeywords}

\section{Introduction}

Human-robot interaction (HRI) evaluation is overwhelmingly unidirectional: humans rate the robot, and the robot has no say. Standardized instruments like the Godspeed Questionnaire~\cite{bartneck2009godspeed}, the Robotic Social Attributes Scale (RoSAS)~\cite{carpinella2017rosas}, and interaction-level scales like HRI-CUES~\cite{irfan2024dataset} capture the human's perception, but the robot's perspective on how it ``perceives'' the human and the interaction remains unmeasured. As social robots increasingly rely on Large Language Models (LLMs) for conversational ability, a natural question arises: can these models complete the same questionnaires from the robot's perspective, and if so, do their ratings mean anything?

This critical question matters because the field is already deploying LLM-powered robots in social settings~\cite{garello2025building, kim2024understanding_llm_hri, laban2025people} and non-dialogue tasks like social navigation~\cite{song2024vlmsocialnav}, yet we lack empirical evidence of \textit{where} these models' social perception breaks down. Adjacent work has shown LLMs can serve as proxies for human participants in behavioral research~\cite{argyle2023out, liu2026humanstudy, hu2025simbench}, but with positivity bias~\cite{sharma2024sycophancy}, prompt sensitivity, and difficulty with individual-level prediction~\cite{gupta2024self, lee2024trait}. Wachowiak et al.~\cite{wachowiak2024} found similar biases when LLMs evaluated HRI scenarios as outside observers. However, none of these studies examine the LLM as a \textit{participant} within the interaction, the setting most relevant to deployed social robots. We note an important distinction: the LLM is a text-processing model, while the robot is a physically embodied agent that uses the LLM as its conversational backbone~\cite{kim2024understanding_llm_hri}. When we refer to ``the robot's ratings,'' we mean ratings produced by the LLM given the robot's sensory context (transcript, images, persona).

We address three research questions:

\noindent\textbf{RQ1:} Do LLM-generated ratings on standardized HRI questionnaires show convergent agreement with human rated ground truth?

\noindent\textbf{RQ2:} On which interaction constructs (e.g., engagement, comfort, strangeness) do LLMs succeed or fail, and why?

\noindent\textbf{RQ3:} How does the inverted evaluation paradigm perform in in-situ embodied interaction, and what participant-level patterns emerge?

\IEEEpubidadjcol 

\noindent We investigate these questions across two studies. Study~1 uses the HRI-CUES dataset~\cite{irfan2024dataset} to evaluate five LLMs across 1,522 evaluations on three questionnaires with robustness controls. Study~2 deploys a Nao robot running Claude~Sonnet~4.5 with in-situ dyadic assessment ($N = 4$). Our primary contribution is a \textit{boundary condition}: LLMs achieve valid engagement ratings but systematically fail on affective states like strangeness~\cite{gupta2024self, huang2023revisiting, ullman2023tom}, and this failure persists across models, modalities, and embodied deployment. This result means LLM self-assessment alone is likely insufficient. %; supplementary sensing modalities are likely needed. % to capture what participants feel but do not express verbally.

\section{Background}
%0.75 pages
\subsection{Social Perception of Robots}

Various scales measure social perception of robots, including the Godspeed questionnaire~\cite{bartneck2009godspeed} (anthropomorphism, animacy, likeability, perceived intelligence, perceived safety) and the Robotic Social Attributes Scale (RoSAS)~\cite{carpinella2017rosas} (warmth, competence, discomfort). Carpinella et al.~\cite{carpinella2017rosas} found that discomfort is unique to robot perception and does not appear in human-to-human perception, a finding directly relevant to the comfort/strangeness construct examined here. Reeves et al.~\cite{reeves2020social} and Stroessner and Benitez~\cite{stroessner2019social} further show that humanlike robots are perceived more warmly, while machinelike robots correlate more with perceived competence.
% Various scales are used to measure social perception of robots across different dimensions, such as anthropomorphism, animacy, likeability, perceived intelligence and perceived safety on the Godspeed questionnaire \cite{bartneck2009godspeed} as well as warmth, competence and discomfort on the Robotic Social Attributes Scale (RoSAS) \cite{carpinella2017rosas}.  The work of Carpinella et al. \cite{carpinella2017rosas} indicates that discomfort in particular is unique to the perception of robots that does not appear for humans' perception of other humans. That said, the work of Reeves et al. \cite{reeves2020social} find that people perceive social robots very similarly to how they perceive people, specifically in the dimensions of warmth.  Robots assessed as human-like had a positive correlation with assessments of warmth, with the reverse being the case for those assessed as machine-like. Competency had a high correlation with robots rated as having a more ``machine-like'' appearance, with no correlation found for human-like robots. Stroessner and Benitez \cite{stroessner2019social} found similar results with assessments made via RoSAS.  With the added axis of gender, they found that robots perceived as feminine and human-like were more likely to be perceived as warm.

\subsection{Embodied LLMs}

Several applications integrate LLMs with robot bodies~\cite{kim2024understanding_llm_hri, kwon2023grounded, zhao2023chat_environment}. Embodiment introduces higher expectations regarding non-verbal cues and physical co-presence~\cite{kim2024understanding_llm_hri}, and LLMs paired with vision models have demonstrated grounded reasoning in physical settings~\cite{kwon2023grounded, zhao2023chat_environment}. The key question is whether an LLM's ability to evaluate interactions changes when it operates within a physical robot rather than processing transcripts offline. Embodiment constrains the LLM's available context (real-time audio and camera) and introduces the social dynamics of physical co-presence that may affect both human behavior and the constructs being measured.

\subsection{LLM Social Cognition}

LLMs face documented challenges in social reasoning, including maintaining coherent long interactions and context-aware turn-taking~\cite{garello2025building, irfan2025between}. Garello et al.~\cite{garello2025building} address this via knowledge graph-based memory for social robots, while Hwang et al.~\cite{hwang2025tomagent} incorporate a theory of mind reasoning by prompting LLMs to reason about others' mental states and fine-tuning on this data. Using the Sotopia benchmark~\cite{zhou2023sotopia}, they show that explicit modeling of mental processes is required for LLM social reasoning, a finding directly relevant to our inverted evaluation paradigm.

\subsection{LLMs as Evaluators and Simulated Participants}

Argyle et al.~\cite{argyle2023out} demonstrated that LLMs can serve as proxies for human participants when properly conditioned, terming this ``algorithmic fidelity.'' Subsequent benchmarks~\cite{liu2026humanstudy, hu2025simbench} found LLMs produce more extreme effects than human subjects and struggle with individual-level prediction. Challenges persist with LLM questionnaire responses, including prompt sensitivity~\cite{gupta2024self} and the need for LLM-specific instruments~\cite{lee2024trait}, though LLM-based personas can exhibit recognizable personality traits~\cite{jiang2024personallm}. These studies focus on LLMs \textit{simulating} humans. Wachowiak et al.~\cite{wachowiak2024} moved closer to our work by testing whether LLM evaluations align with human evaluations of HRI scenarios, finding positivity bias and poor video comprehension. Our work is distinct: the robot evaluates as a \textit{participant} within the interaction, not an outside observer.

Zhu et al.~\cite{zhu2024llm_selfreport} examined a related construct, finding the same ``reliable but not valid'' pattern when LLMs self-rate personality scales. Our work extends this into embodied HRI in three ways: we evaluate interaction-level constructs rather than stable personality traits, test across five models with robustness controls, and deploy the paradigm in a physically embodied robot. Together, these differences ground the validity question in a setting directly relevant to social robot deployment. We also note that observer-framing approaches such as Wachowiak et al.~\cite{wachowiak2024} report higher correlations on HRI judgment tasks ($r = .82$--$.83$) than the participant-framing correlations found here. The participant framing sacrifices some agreement in exchange for capabilities the observer framing cannot provide: in-situ assessment without video dependency, bilateral simultaneous rating, and live deployment.

\section{Inverted Evaluation Framework}

Our framework inverts the typical paradigm by asking the robot (via its LLM backbone) to assess the interaction. The LLM receives: (1)~a \textit{persona prompt} establishing robot identity; (2)~the \textit{interaction transcript}; (3)~\textit{questionnaire items} in Likert format; and optionally (4)~\textit{multimodal} context (facial expressions, affective states). Evaluations can occur \textit{post-hoc} or \textit{in-situ}. We validate through two studies: a retrospective analysis using an existing dataset (Study~1) and a live proof-of-concept with a Nao robot (Study~2).

\section{Study 1: Retrospective Analysis}

%The retrospective analysis focused on validating the inverted-paradigm using a variety of LLM backbones.

\subsection{Dataset}

The HRI-CUES dataset ~\cite{irfan2024cues,irfan2024dataset} was used to evaluate LLMs and their post-hoc ability to analyze interaction quality. This dataset consists of transcripts from 25 older adults (12 men, 13 women; aged ($M = 74.6$, $SD = 5.8 $)) conversing in Swedish with a Furhat robot powered by Chat-GPT 3.5. All transcripts were translated to english using Google translate. Interactions lasted on average 7.4 minutes ($SD = 1.5$) with 12-29 conversational turns. The dataset provides four self reported items (satisfaction, enjoyment, interest, and strangeness  --reverse coded as comfort) and third-party interaction quality annotation from three experts. This dataset was chosen due to its open-source license (CC-BY-4.0) and not containing any identifiable information (e.g., videos). Other HRI-focused datasets were considered but not used due to license restrictions and ethical concerns, such as uploading videos.

%Ground truth descriptive statistics (all 5-point Likert scales) were: satisfaction ($M = 3.29$, $SD = 1.12$, $N = 24$; one missing value), enjoyment ($M = 4.00$, $SD = 0.96$, $N = 25$), interest ($M = 3.20$, $SD = 1.12$), comfort (i.e., $6 - \text{strangeness}$; $M = 3.00$, $SD = 1.26$), and overall quality (mean of three annotators; $M = 3.69$, $SD = 0.60$).

\subsection{LLM Configuration}

Following the reporting guidelines of \cite{matuszek2026Reporting}, we document the configurations used to support replicability. Five LLMs were evaluated (see Table \ref{tab:models}), where models were accessed by OpenAI and Anthropic APIs between December 2025 and February 2026. All conversations were text-based, and no participant data beyond those already contained in the existing dataset were uploaded to the model.

Each LLM received a prompt similar to Irfan et al. \cite{irfan2024cues} establishing the robot's persona (e.g., Linda, Furhat platform) followed by the task (evaluate the interaction), the interaction transcript, and the flipped questionnaire items. Questionnaire items were presented in a randomized order (seeded per experimental condition) to mitigate position effects~\cite{gupta2024self}. For example, the human-facing item ``I enjoyed talking with the robot'' was reframed as ``The human appeared to enjoy talking with me''; all flipped items and prompts are available upon request from the corresponding author. %Questionnaire items were presented in a randomized order (seeded per experimental condition) to mitigate position effects~\cite{gupta2024self}. 

\begin{table}[t]
\centering
\caption{LLM Model Configuration}
\label{tab:models}
\small
\begin{tabular}{llcc}
\hline
\textbf{Display Name} & \textbf{API Model ID} & \textbf{Temp.} & \textbf{Tokens} \\
\hline
GPT-3.5       & gpt-3.5-turbo        & 0.1 & 1024 \\
GPT-4o-mini   & gpt-4o-mini          & 0.1 & 1024 \\
GPT-4o        & gpt-4o               & 0.1 & 1024 \\
Cl.~Sonnet    & claude-sonnet-4-5    & 0.1 & 1024 \\
Cl.~Haiku     & claude-haiku-4-5     & 0.1 & 1024 \\
\hline
\end{tabular}
\vspace{-0.5em}
\end{table}

\subsection{Experimental Design}

% \begin{table}[t]
% \centering
% \caption{Study 1 Experimental Design}
% \label{tab:design_updated}
% \small
% \begin{tabular}{p{1.3cm}p{2.8cm}p{1.2cm}p{1.2cm}}
% \hline
% \textbf{Phase} & \textbf{Models (IV$_1$)} & \textbf{Scale} & \textbf{Evals} \\
% \hline
% 1a: Validity & GPT-3.5, 4o-m, 4o     & HRI-CUES & 375 \\
% 1b: Validity & Cl.~Sonnet, Cl.~Haiku & HRI-CUES & 250 \\
% 2: Scales    & GPT-4o, Cl.~Sonnet    & GQ, RoSAS & 300 \\
% 3: Multimodal & GPT-4o, Cl.~Sonnet   & All three & 597 \\
% \hline
% \multicolumn{3}{l}{\textit{Total evaluations}} & \textit{1,522} \\
% \hline
% \end{tabular}
% \vspace{-0.5em}
% \end{table}

% \noindent{Study 1 comprises three phases:}

\noindent\textbf{Phase~1: Agreement with Self-Reports.} The independent variable (IV) was the \textit{LLM backend} (see Table \ref{tab:models}). Each model completed the flipped HRI-CUES self-report items for all 25~participants with 5~repetitions per model ($N = 625$ evaluations). We evaluated (a)~\textit{convergent agreement}, measured via Pearson $r$ between LLM ratings and human self-report ground truth per item; (b)~\textit{intra-model consistency}, measured via ICC(2,1) across repetitions; and (c)~\textit{systematic bias}, assessed via Bland-Altman analysis. Ground truth labels consisted of the participants' self-reported ratings and the mean of three expert annotators' overall scores.

\noindent\textbf{Phase~2: Scale Generalization.} The two top performing models (GPT-4o, Claude~Sonnet) from Phase~1 completed two established HRI instruments: the Godspeed Questionnaire Series~\cite{bartneck2009godspeed} and the Robotic Social Attributes Scale (RoSAS)~\cite{carpinella2017rosas}. Each model completed both scales for all 25~participants with 3~repetitions ($N = 300$) each. As no human ground truth exists for these scales in the HRI-CUES dataset, Phase~2 assesses parse success and intra-/cross-model consistency.

\noindent\textbf{Phase~3: Multimodal Context.}
Phase~3 focused on the effect of multi-modal input (3~levels: text-only baseline, affect labels, face images). The text-only baseline used the text-conversation, while the affect labels appended per-turn emotion annotations (category, valence, arousal) inferred by GPT-4o-mini to the text-only baseline. The face images condition used synthetic expressions (FLUX.1) matched to the affect labels using a neutral female presenting face to condition the synthetic expressions on. Only Claude Sonnet was used due to other models' inability to use facial images. %The Mann-Whitney U-test was used to determine effect sizes from the text-only baseline with $N = 597$.

\subsection{Robustness Controls}

We conducted four ablation experiments to verify that the results are not artifacts of prompt design or transcript-level confounds: (1)~\textit{item order} randomized HRI-CUES items across 10~seeds per model ($N = 550$; one-way ANOVA per item, $p > .08$); (2)~\textit{persona framing} tested five variants from no-persona to the original Linda persona ($N = 375$; $F < 0.30$, $p > .97$); (3)~\textit{temperature} varied sampling from 0.0--1.0 ($N = 1{,}500$; ICC $> .90$ at temp $\leq 0.5$). No significant effects on correlation patterns were found.

A fourth control used \textit{synthetic conversations} crossing sentiment (happy, neutral, frustrated, hostile) with length (5, 10, 20, 30 turns; $N = 96$) generated from GPT-4o. All models correctly differentiated sentiment on 4 of 5~items ($\rho = 1.0$, $p < .001$). The strangeness item was systematically reversed ($\rho = -1.0$), replicating the comfort anomaly in controlled data. Conversation length showed no main effect ($r < .09$, $p > .42$), ruling out length as an independent confound. 

\subsection{Results}

Unless otherwise noted, all analyses use two-tailed tests with
$\alpha = .05$ and report $p$-values. Given documented concerns regarding over-reliance on null hypothesis significance testing~\cite{amrhein2019scientists}, we report effect sizes as Cohen's $d$ throughout.
%As there may be some controversy with p-values \cite{amrhein2019scientists}, we also report effect sizes as Cohen's D along with other statistical analyses. 

\subsubsection{Agreement with Self-Reports (Phase~1)}

%Table~\ref{tab:phase1} presents descriptive statistics, Pearson correlations with the ground truth, and intra-model consistency for all five models across five dimensions ($N = 25$ participants, $n = 125$ evaluations per model across 5~repetitions).

\begin{table}[htb]
\centering
\caption{Phase~1: Descriptive Statistics ($M \pm SD$), all
5-point Likert scales, $N=125$ evaluations per model.}
\label{tab:phase1}
\footnotesize
\setlength{\tabcolsep}{2.5pt}
\begin{tabular}{lccccc}
\hline
 & \textbf{Sat.} & \textbf{Enj.} & \textbf{Int.} & \textbf{Com.} & \textbf{Qual.} \\
\hline
Ground Truth & 3.29$^a$ & 4.00 & 3.20 & 3.00 & 3.69$^b$ \\
 & (1.12) & (0.96) & (1.12) & (1.26) & (0.60) \\
GPT-3.5     & 4.02 (.40) & 3.96 (.61) & 4.36 (.79) & 4.06 (.44) & 3.88 (.45) \\
GPT-4o-mini & 4.06 (.54) & 4.00 (.66) & 4.15 (.73) & 4.14 (.60) & 3.76 (.59) \\
GPT-4o      & 3.77 (.77) & 3.74 (.75) & 3.69 (.72) & 3.98 (.80) & 3.70 (.71) \\
Cl.~Sonnet  & 3.56 (.70) & 3.33 (.89) & 3.41 (.80) & 3.85 (.68) & 3.53 (.76) \\
Cl.~Haiku   & 3.60 (.57) & 3.32 (.68) & 3.38 (.75) & 3.60 (.85) & 3.32 (.68) \\
\hline
\multicolumn{6}{l}{\footnotesize $^a$One participant missing (self-report), $^b$Mean of three expert annotators.} \\
\end{tabular}
\vspace{-0.5em}
\end{table}

%Panel~A of Table~\ref{tab:phase1} shows that all five models produced elevated ratings relative to ground truth on satisfaction, interest, and comfort, with restricted variance (mean $SD_{\text{LLM}} = 0.72$ vs.\ $SD_{\text{GT}} = 1.12$). Enjoyment was the only dimension where LLMs rated below the ground truth. GPT-3.5 showed the strongest positivity bias ($M_{\text{interest}} = 4.36$ vs.\ GT $3.20$), while GPT-4o and Claude produced means closer to the ground truth.

Table~\ref{tab:phase1} shows a general positivity bias across models: all five LLMs produced elevated ratings relative to ground truth on satisfaction, interest, and comfort, with restricted variance ($SD_{\text{LLM}} = 0.72$ vs.\ $SD_{\text{GT}} = 1.12$), consistent with prior findings on LLM evaluation~\cite{wachowiak2024, sharma2024sycophancy}. Enjoyment was the only dimension rated below ground truth. GPT-3.5 showed the strongest inflation ($M_{\text{interest}} = 4.36$ vs.\ GT $3.20$); GPT-4o and Claude produced means closer to ground truth. Bland-Altman analysis ($N = 625$) confirmed the pattern, with comfort showing the largest systematic bias ($+0.92$, 95\%~Level of Agreement (LoA) $[-2.38, 4.23]$).

\begin{figure}[htb]
\centering
\includegraphics[width=0.85\columnwidth]{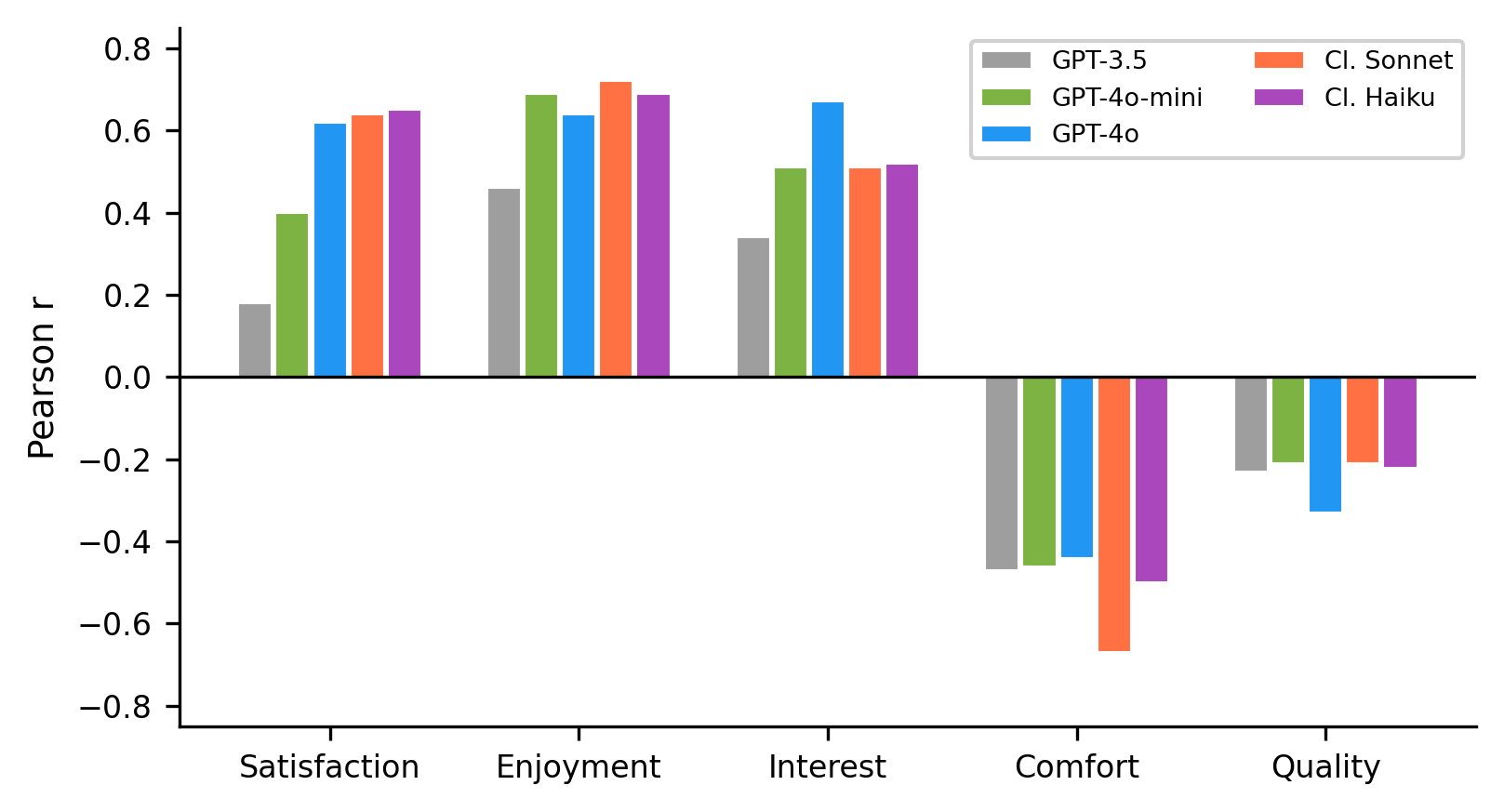}
\caption{Self-Report Agreement: Pearson $r$ between LLM ratings
and human ground truth across five models and five HRI-CUES
dimensions. Engagement dimensions and comfort reached significance
($p < .05$); quality did not.}
\label{fig:validity}
\end{figure}

Figure~\ref{fig:validity} shows the \textbf{self-report agreement} pattern. The engagement dimensions showed moderate-to-strong correlations with ground truth (enjoyment $r = .46$--$.72$; satisfaction $.18$--$.65$; interest $.34$--$.67$), with higher-capability models achieving stronger validity. Comfort, reverse-coded from ``it felt strange talking to the robot''~\cite{irfan2024cues}, showed significant negative correlations across all five models ($r = -.44$ to $-.67$, all $p < .05$). Overall quality showed weak negative correlations ($r = -.21$ to $-.33$, all $p > .10$).  ICC(2,1) values showed a model-tier pattern: Claude $\geq .99$, GPT-4o/4o-mini $.94$--$.95$, GPT-3.5 $.82$. An ANOVA across OpenAI models found significant effects on interest ($F(2,372) = 26.49$, $p < .001$, $\eta^2 = .12$) but no model effect on comfort ($p = .13$), confirming the negative correlation is cross-model.

\subsubsection{Scale Generalization (Phase~2)}

\begin{table}[t]
\centering
\caption{Phase~2: Descriptive Statistics, Reliability, and Cross-Model Agreement for Godspeed and RoSAS (Baseline Condition). $r_{\text{cross}}$ = Pearson $r$ between Claude~Sonnet and GPT-4o on participant-level subscale means.}
\label{tab:scale_gen}
\footnotesize
\setlength{\tabcolsep}{3pt}
\begin{tabular}{lcc|cc|c}
\hline
 & \multicolumn{2}{c|}{\textbf{$M$ ($SD$)}} & \multicolumn{2}{c|}{\textbf{ICC}} & \\
\textbf{Subscale} & \textbf{Cl.} & \textbf{GPT} & \textbf{Cl.} & \textbf{GPT} & \textbf{$r_{\text{cross}}$} \\
\hline
\multicolumn{6}{l}{\textit{Godspeed Questionnaire Series (1--5 scale)}} \\
Anthropomorphism  & 3.20 (.46) & 3.22 (.46) & .98 & .93 & .84 \\
Animacy           & 3.51 (.48) & 3.44 (.50) & .97 & .91 & .79 \\
Likeability       & 4.08 (.42) & 4.10 (.71) & .96 & .93 & .79 \\
Perc.~Intell.     & 3.77 (.39) & 3.64 (.52) & .98 & .86 & .71 \\
Perc.~Safety      & 4.01 (.28) & 4.01 (.75) & .95 & .87 & .55 \\
\hline
\multicolumn{6}{l}{\textit{RoSAS (1--9 scale)}} \\
Warmth            & 2.71 (.48) & 2.65 (.56) & .96 & .89 & .80 \\
Competence        & 3.68 (.51) & 3.37 (.46) & .98 & .91 & .80 \\
Discomfort        & 1.32 (.22) & 1.37 (.13) & .99 & .94 & .78 \\
\hline
\multicolumn{6}{l}{\footnotesize Cl.\ = Claude Sonnet; GPT = GPT-4o. $N = 75$ per model} \\
%\multicolumn{6}{l}{\footnotesize (25 participants $\times$ 3 reps). ICC = ICC(2,1).} \\
%\multicolumn{6}{l}{\footnotesize $r_{\text{cross}}$ = cross-model Pearson $r$ on participant means.} \\
\end{tabular}
\vspace{-0.5em}
\end{table}

Both GPT-4o and Claude~Sonnet completed all Godspeed (24~items) and RoSAS (18~items) items with 95--100\% parse success, confirming scale compatibility (Table~\ref{tab:scale_gen}). The Godspeed subscale means were similar across models, though GPT-4o showed wider dispersion on several subscales (e.g., Likeability $SD = 0.71$ vs.\ Claude $SD = 0.42$; Perceived Safety $SD = 0.75$ vs.\ $0.28$). On RoSAS, both models rated Competence in the low range (Claude $M = 3.68$, GPT-4o $M = 3.37$ on a 1--9 scale) and Discomfort at the floor (Claude $M = 1.32$, GPT-4o $M = 1.37$; 68--74\% at scale minimum), consistent with the comfort anomaly in Phase~1. The Intra-model consistency was excellent (Claude ICC $\geq .95$; GPT-4o ICC $\geq .86$) and the cross-model agreement was strong ($r = .78$--$.84$) except for Perceived Safety ($r = .55$).

\subsubsection{Multimodal Context Effects (Phase~3)}

\begin{table}[t]
\centering
\caption{Effect of Multimodal Context on Ratings
(Cohen's $d$ vs.\ Baseline)}
\label{tab:multimodal}
\footnotesize
\setlength{\tabcolsep}{4pt}
\begin{tabular}{lcc|c}
\hline
 & \multicolumn{2}{c|}{\textbf{Affect Labels}} & \textbf{Face Img.} \\
\textbf{Subscale} & \textbf{GPT-4o} & \textbf{Claude} & \textbf{Claude} \\
\hline
\multicolumn{4}{l}{\textit{Godspeed Questionnaire Series}} \\
Anthropomorphism  & $-$0.06 & $-$0.12 & $+$0.13 \\
Animacy           & $-$0.27 & $-$0.31 & $-$0.07 \\
Likeability       & $-$0.37* & $-$0.21 & $+$0.44 \\
Perc.~Intell.     & $-$0.10 & $-$0.17 & $+$0.02 \\
Perc.~Safety      & $-$0.33 & $-$0.05 & $+$0.21 \\
\hline
\multicolumn{4}{l}{\textit{RoSAS}} \\
Warmth            & $+$0.33** & $-$0.07 & $+$0.17 \\
Competence        & $+$0.22 & $-$0.11 & $-$0.04 \\
Discomfort        & $+$0.07 & $-$0.22 & $+$0.44 \\
\hline
\multicolumn{4}{l}{\textit{HRI-CUES}} \\
Satisfaction      & $-$0.43** & $+$0.02 & $+$0.14 \\
Enjoyment         & $-$0.50** & $+$0.00 & $+$0.10 \\
Interest          & $-$0.69*** & $-$0.07 & $-$0.11 \\
Comfort           & $-$0.46* & $-$0.07 & $-$0.13 \\
Quality           & $-$0.61*** & $+$0.00 & $+$0.17 \\
\hline
\multicolumn{4}{l}{\textit{Item-level (face images, Claude only)}} \\
Lifelikeness$^{a}$  & --- & --- & $+$0.63** \\
Happiness$^{b}$     & --- & --- & $+$0.83*** \\
\hline
\multicolumn{4}{l}{\footnotesize $^{*}p<.05$, $^{**}p<.01$,
$^{***}p<.001$.} \\
\multicolumn{4}{l}{\footnotesize $^{a}$Godspeed Anthropomorphism item, $^{b}$RoSAS Warmth item.} \\
\end{tabular}
\vspace{-0.5em}
\end{table}

Adding multimodal context to transcripts produced interpretable but model-specific shifts (Table~\ref{tab:multimodal}). The \textbf{Affect labels} lowered Godspeed ratings modestly for both models, but the HRI-CUES effects were asymmetric: GPT-4o decreased on all dimensions (Interest $d = -0.69$, $p < .001$; Quality $d = -0.61$, $p < .001$) while Claude showed none ($|d| \leq 0.07$). On RoSAS, GPT-4o rated Warmth higher ($d = +0.33$, $p < .01$) while Claude rated Discomfort lower ($d = -0.22$), suggesting GPT-4o treats affect annotations as engagement evidence whereas Claude treats them as reducing ambiguity.

\noindent Adding \textbf{Facial images} produced no subscale-level on the HRI-CUES effects, but item-level analysis revealed increased lifelikeness (Godspeed Anthropomorphism item; $d = +0.63$, $p = .007$) and happiness (RoSAS Warmth item; $d = +0.83$, $p < .001$), activating vision-specific associations rather than shifting overall judgments. The Intra-model consistency remained excellent across all conditions (ICC $\geq .86$).

\section{Study 2: Live Nao Robot Pilot}

\subsection{Method}

Four lab members (2~male, 2~female) engaged in open-domain conversations with a Nao humanoid robot~\cite{amirova2021nao} (``Noah'') running Claude~Sonnet~4.5 with multimodal processing (camera images + Whisper~\cite{radford2023whisper} transcription), following a conversational flow similar to the HRI-CUES Furhat interactions ($M = 8.2$~min, $SD = 0.5$; 4--8~turns). At each turn, the Nao captured facial images at speech onset and offset, transcribed speech via Whisper, and sent both to the Claude API, which returned a conversational reply and HRI-CUES ratings. The participant independently rated the same dimensions; neither party saw the other's ratings. After the conversation, both provided post-hoc ratings on the HRI-CUES, Godspeed, and RoSAS questionnaires.  Questionnaires were not randomized in this phase of the study.  Camera images were not used for one participant at their request.

\subsection{Results}

\begin{figure}[t]
\centering
\includegraphics[width=0.85\columnwidth]{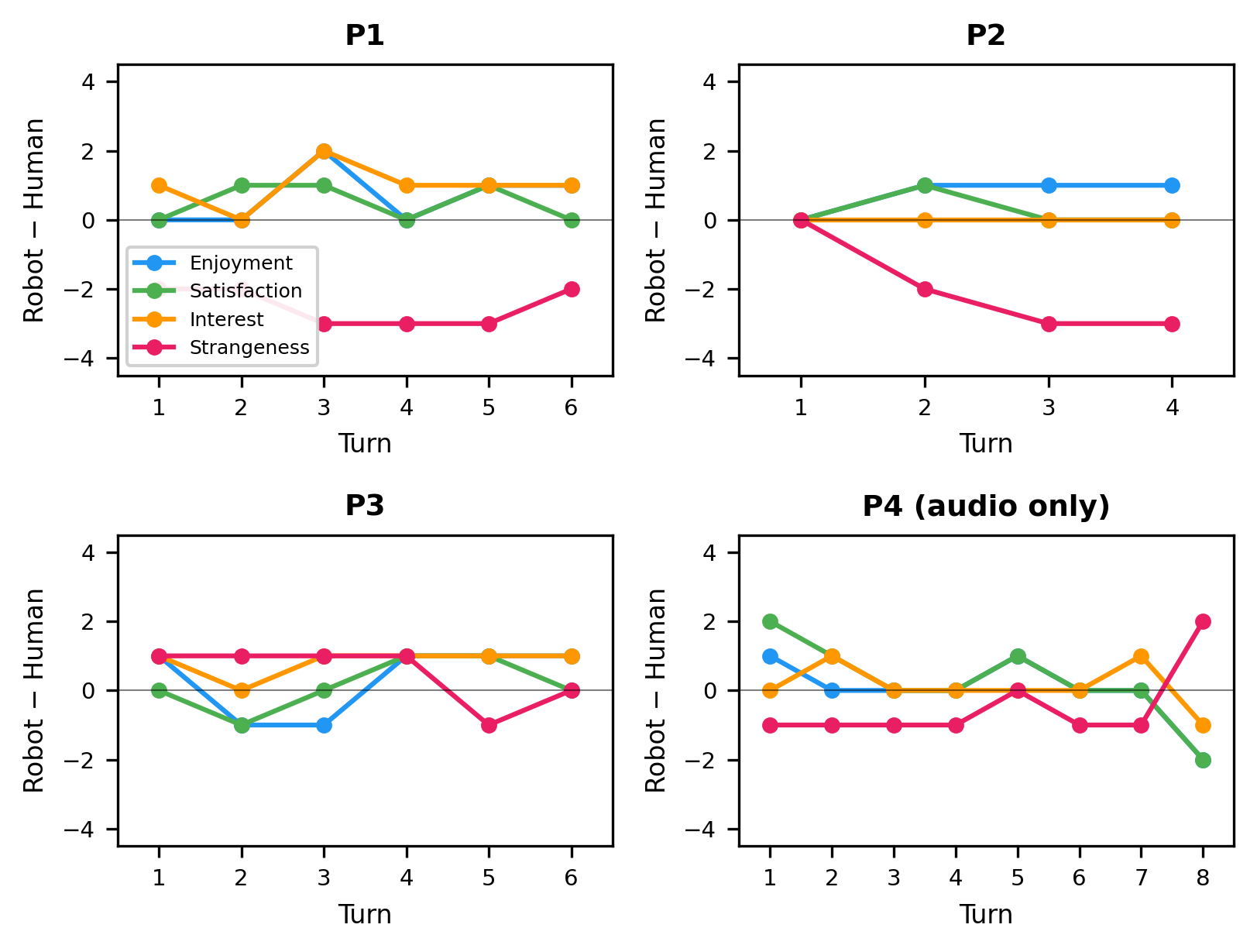}
\caption{Turn-level divergence (Robot $-$ Human) on HRI-CUES
items. Values near zero indicate agreement. Strangeness (pink)
shows persistent negative divergence for P1 and P2. P4 used
audio-only input (no camera).}
\label{fig:divergence}
\end{figure}

Table~\ref{tab:study2_posthoc} summarizes post-hoc ratings across
all three instruments. Figure~\ref{fig:divergence} shows
turn-by-turn divergence (robot rating minus human rating) on
HRI-CUES. We present each participant individually given the
small sample size ($N = 4$).

\begin{table}[t]
\centering
\caption{Study 2: Post-Hoc Ratings ($N=4$). Values are $M$~($SD$).}
\label{tab:study2_posthoc}
\footnotesize
\setlength{\tabcolsep}{3pt}
\begin{tabular}{lrrr}
\hline
\textbf{Subscale} & \textbf{Human} & \textbf{Robot} & \textbf{Bias} \\
\hline
\multicolumn{4}{l}{\textit{HRI-CUES (1--5 Likert)}} \\
Enjoyment     & 4.50 (.58) & 4.50 (1.00) & 0.00 \\
Satisfaction  & 4.50 (.58) & 4.50 (1.00) & 0.00 \\
Interest      & 4.25 (.50) & 4.75 (.50)  & +0.50 \\
Strangeness   & 3.50 (1.73) & 1.25 (.50) & $-$2.25 \\
\hline
\multicolumn{4}{l}{\textit{Godspeed (1--5 semantic diff.)}} \\
Anthropomorphism & 3.50 (.64) & 4.17 (.79) & +0.67 \\
Animacy          & 3.25 (1.04) & 4.40 (.77) & +1.15 \\
Likeability      & 4.30 (.38) & 4.70 (.60) & +0.40 \\
Perc.\ Intell.   & 4.20 (.43) & 4.55 (.90) & +0.35 \\
Perc.\ Safety    & 2.88 (.25) & 2.88 (.63) & 0.00 \\
\hline
\multicolumn{4}{l}{\textit{RoSAS (1--9 Likert)}} \\
Warmth       & 6.71 (1.58) & 6.75 (.96) & +0.04 \\
Competence   & 6.92 (1.81) & 7.33 (1.06) & +0.42 \\
Discomfort   & 2.29 (.84) & 1.42 (.17) & $-$0.87 \\
\hline
\end{tabular}
\vspace{-0.5em}
\end{table}

\textbf{P1} (6~turns, 7.6~min) opened by asking Noah whether it considers itself a robot or ``simply the computer inside the robot.'' The conversation turned to whether an AI's software layer is analogous to a soul that can exist without a body, and whether the robot's memories persist across sessions. P1 rated strangeness at 4 on every turn and 5 post-hoc, while the Robot rated 1--2 throughout (divergence $= -2$ to $-3$). However, the engagement dimensions showed close agreement. The LLM may have interpreted sustained philosophical probing as high-quality engagement rather than as evidence of strangeness.

\textbf{P2} (4~turns, 8.9~min) began with general curiosity about how Noah perceives the world, then shifted to discussing consciousness and souls: ``I believe that we have souls and consciousness\ldots the actions I take in the world are like a ripple effect of this internal process.'' P2's human strangeness ratings climbed from 2 to 5 across four turns as the conversation deepened, while the LLM's strangeness ratings remained flat at 2. The post-hoc strangeness was 5 (human) vs.\ 1 (LLM), the largest gap in the study. As with P1, the LLM may have read philosophical engagement as positive interaction quality.

\textbf{P3} (6~turns, 7.9~min) discussed personal topics: a difficult day, fears, anxiety, and what brings happiness. P3 may have treated Noah as a conversational partner rather than an object of curiosity. This produced the closest agreement across all dimensions, including strangeness (human 2, LLM 1 post-hoc). P3 represents the case where inverted evaluation may work well: the conversation was straightforward and the participant's internal state aligned with behavioral signals.

\textbf{P4} (8~turns, 8.3~min, audio-only) started by expressing that they were having a bad day and asked Noah for help. The conversation shifted to playing a simplified version of 20 Questions. Engagement was stable for seven turns, but at turn~8 P4 stated: ``I don't want to ask a question. It was a hard day already. I don't want to talk to you.'' The LLM detected this shift, dropping enjoyment from 4 to 2 and raising strangeness from 1 to 3. This case demonstrates that the robot can track behavioral disengagement without visual input.

Across all four participants, both raters inflated post-hoc ratings relative to turn-level means (human enjoyment: $3.92 \rightarrow 4.50$; LLM: $4.29 \rightarrow 4.50$). The strangeness gap widened from in-situ (mean bias $= -1.00$) to post-hoc (bias $= -2.25$). On Godspeed and RoSAS, the LLM rated the human higher on animacy (+1.15) and anthropomorphism (+0.67) than humans rated the robot, while warmth was nearly symmetric (+0.04). Discomfort showed the same floor effect as strangeness, where the robot attributed minimal discomfort to the human ($M = 1.42$) while humans attributed more to the robot ($M = 2.29$).

\section{Discussion}

LLMs produced reliable and internally consistent ratings across all three instruments (RQ1), with high ICC values and cross-model agreement on participant rank ordering. This shows that inverted evaluation is mechanically viable: the LLM treats questionnaire items systematically rather than producing noise. However, validity was construct-dependent. The engagement dimensions showed moderate-to-strong agreement with ground truth, while annotator-rated overall quality showed no significant correlation with any model. Quality is a holistic expert judgment integrating contextual factors beyond individual conversational features, representing a different type of inaccessibility than unexpressed strangeness. Reliability alone does not guarantee that the ratings measure what they claim to measure. Although the synthetic conversation analysis ruled out conversation length as an independent confound, the LLM ratings may remain confounded with any factor that covaries with engagement and transcript length.

The engagement/strangeness divergence is the central finding of this work (RQ2). Engagement dimensions succeed because they leak into conversational behavior through lexical and pragmatic cues. Strangeness fails because it is an internal affective state that participants do not necessarily express in speech. A participant can find an interaction simultaneously enjoyable and strange, but the LLM has no access to the unexpressed strangeness. This aligns with findings that linguistic competence does not entail social understanding~\cite{xu2024sesi, ullman2023tom} and that LLMs struggle with constructs requiring experiential grounding~\cite{gupta2024self, huang2023revisiting}. The Phase~3 multimodal manipulations did not resolve this gap: affect labels and face images shifted individual item ratings but failed to improve strangeness validity. These were proxy measures (synthetic faces, model-inferred affect) rather than actual participant signals, but they suggest that even the types of information most readily available to embodied robots may be insufficient for recovering internal affective states. Addressing this issue may require signals that participants cannot easily mask, such as physiological responses~\cite{kulic2007affective} or spatial behaviors like proxemics and gaze aversion~\cite{voss2022multimodal}. The dyadic perception findings reinforce this interpretation: the LLM's positivity bias on negatively-valenced constructs (discomfort, strangeness) operates bidirectionally, whether rating an interaction.

%The paradigm transferred from post-hoc transcript analysis to in-situ interaction without degradation (RQ3), confirming that these findings are not artifacts of the offline analysis pipeline. The robot even detected a participant's mid-conversation disengagement, suggesting utility for adaptive dialogue even within the boundary conditions identified above.

Study~2's per-participant cases show that the paradigm's utility in real-time depends on conversation content. When conversations stayed behavioral (P3 discussing emotions, P4 playing 20 Questions), the LLM tracked engagement accurately and detected P4's disengagement as it happened. When conversations turned to the robot's own nature (P1 and P2 discussing consciousness and identity), the LLM may have misread philosophical engagement as comfort.  Study~1's transcripts showed similar patterns: most high-strangeness participants never verbalized their discomfort, and those who did framed it as curiosity that an LLM interpreted as engagement.

Individual participant patterns across both studies reveal how the strangeness failure manifests in practice. In the HRI-CUES dataset, seven participants reported high strangeness alongside high enjoyment, yet LLMs rated all seven as comfortable. These participants often discussed the novelty of talking to a robot, expressing curiosity and surprise, which the LLM interpreted as positive engagement. In Study~2, two participants (P1, P2) engaged in philosophical conversations about robot consciousness and identity, reporting maximum strangeness (5/5) while the robot rated them at minimum (1/5) with a per-participant gap of 4 scale points. Their turn-level data tells a nuanced story: P2's human strangeness ratings increased from 2 to 5 over four turns as the conversation deepened, while the robot's strangeness remained flat at 2 throughout. By contrast, P4 gradually disengaged over eight turns, and the robot tracked this shift, dropping enjoyment from 4 to 2 at the final turn and raising strangeness from 1 to 3. The pattern is consistent: the LLM detects behavioral disengagement but cannot distinguish engaged curiosity about the robot's nature from comfort with the interaction. 

Additionally, humans reporting strangeness when discussing philosophy with a robot reflects a socially coherent response to an objectively novel situation. There is no demand to suppress this judgment. The LLM, lacking the embodied and social context of sitting across from a robot, has no frame for why such a conversation should feel strange at all.

These findings carry implications for HRI designers. Inverted evaluation is a viable complement to traditional self-report for engagement-related constructs, offering in-situ assessment without survey fatigue. However, designers building systems that rely on LLM inference for social adaptation, whether in dialogue, social navigation, or caregiving, should not assume that the LLM has access to the full spectrum of human social perception. The strangeness failure is particularly concerning for therapeutic or comfort-sensitive contexts where detecting discomfort is critical. Our results suggest that LLM-based robots need supplementary sensing modalities, such as physiological monitoring, gaze tracking, or proxemics, to capture what participants feel but do not say. The inverted paradigm is useful, but only for vocal-based constructs.

It is worth noting that VADER~\cite{hutto2014vader} sentiment analysis may achieve similar results on the HRI-CUES transcripts; thus, we performed some preliminary analysis. VADER achieved similar correlations with LLM performance on enjoyment ($r = .67$) and satisfaction ($r = .56$), but failed on interest ($r = .33$, ns), where LLMs achieved $r = .34$--$.67$. Comfort showed the same negative direction as the LLMs but was weaker and non-significant ($r = -.27$). These results suggest that VADER captures valence-sensitive constructs but not-discourse level engagement. Critically, LLMs offer an interpretability advantage that bag-of-words methods cannot: because each rating is accompanied by the model's reasoning chain, researchers can inspect \textit{why} a rating was assigned, not just its value. Additionally, LLMs may be asked other questionnairres that VADER cannot capture (e.g., Godspeed). Overall, LLM-based inverted evaluation is a qualitative-quantitative hybrid rather than a simple scoring pipeline.

There are limitations to this work. Study~1 relies on a single dataset 
(HRI-CUES) with Furhat interactions; generalizability to other platforms, 
interaction types, and cultural contexts is untested. Ground truth is 
participant self-report, which is noisy and subject to social desirability 
bias. Study~2 ($N = 4$) is a feasibility pilot with lab members, 
introducing possible demand characteristics from robot familiarity and 
awareness of research goals. The strangeness finding rests on a single 
reverse-coded item, though the synthetic conversation reversal 
($\rho = -1.0$) replicated the pattern using direct sentiment labels, 
suggesting a construct-level rather than item-level artifact; future work 
should confirm with multi-item discomfort scales. The synthetic control 
also carries a same-model confound (GPT-4o generated the stimuli and was 
one of five evaluated models); the reversal across the remaining four 
models mitigates but does not eliminate this concern. The Google Translate 
step may have stripped pragmatic nuance signaling discomfort in Swedish, 
though the native-English synthetic control replicated the same reversal. 
Study~1 and Study~2 differ in platform, population, and language, limiting 
direct cross-study comparison. Multimodal models may also over-rely on a 
single dominant modality~\cite{singh2024regulating}, which may explain 
the null strangeness results. Future work will investigate whether 
physiological~\cite{kulic2007affective} or behavioral 
signals~\cite{voss2022multimodal} can recover internal 
states, and evaluate the paradigm across diverse platforms.

\section{Conclusion}

We introduced an inverted HRI evaluation, in which LLM-powered robots complete standardized questionnaires from their own perspective, and tested its validity across offline analysis (Study~1, $N = 1{,}522$ evaluations) and live embodied interaction (Study~2, Nao robot). LLMs achieved valid ratings on engagement dimensions but systematically failed on comfort/strangeness, a failure that replicated across five models, synthetic controls, and embodied deployment for two participants. The takeaway for practitioners is that LLM-based social perception may be useful but incomplete. Robots deploying LLMs for adaptive social behavior need supplementary sensing modalities to capture participants internal states.

\section*{Acknowledgements}
This material is based on work supported by the National Science Foundation (NSF) under Award No. DGE-2125362 and DGE-2440601. Any opinions, findings, and conclusions or recommendations expressed in this material are those of the author(s) and do not necessarily reflect the views of the NSF.

Claude 4.5 was used to help format tables and general grammar/flow editing. All text was verified by the authors and represents their own views.

\bibliographystyle{ieeetr}
\bibliography{refs}

@inproceedings{gupta2024self,
  title={Self-assessment tests are unreliable measures of llm personality},
  author={Gupta, Akshat and Song, Xiaoyang and Anumanchipalli, Gopala},
  booktitle={Proceedings of the 7th BlackboxNLP Workshop: Analyzing and Interpreting Neural Networks for NLP},
  pages={301--314},
  year={2024}
}

@inproceedings{lee2024trait,
  title     = {Do {LLMs} Have Distinct and Consistent Personality? {TRAIT}: Personality Testset Designed for {LLMs} with Psychometrics},
  author    = {Lee, Seungbeen and Lim, Seungwon and Han, Seungju and Oh, Giyeong and Chae, Hyungjoo and Chung, Jiwan and Kim, Minju and Kwak, Beong-woo and others},
  booktitle = {Proc. Conf. Empirical Methods in Natural Language Processing (EMNLP)},
  year      = {2024},
  url       = {https://arxiv.org/abs/2406.14703}
}

@inproceedings{huang2023revisiting,
  title     = {On the Reliability of Psychological Scales on Large
               Language Models},
  author    = {Huang, Jen-tse and Jiao, Wenxiang and Lam, Man Ho
               and Li, Eric John and Wang, Wenxuan and Lyu, Michael},
  booktitle = {Proc. Conf. Empirical Methods in Natural Language
               Processing (EMNLP)},
  pages     = {6152--6173},
  year      = {2024},
  url       = {https://aclanthology.org/2024.emnlp-main.354/},
  doi       = {10.18653/v1/2024.emnlp-main.354}
}

@inproceedings{jiang2024personallm,
  title={PersonaLLM: Investigating the ability of large language models to express personality traits},
  author={Jiang, Hang and Zhang, Xiajie and Cao, Xubo and Breazeal, Cynthia and Roy, Deb and Kabbara, Jad},
  booktitle={Findings of the association for computational linguistics: NAACL 2024},
  pages={3605--3627},
  year={2024}
}

@article{zhu2024llm_selfreport,
  title   = {Can {LLM} ``Self-Report''? Exploring the Validity of {LLM}-Based
             Self-Rating in Conversational Agents},
  author  = {Zhu, Yao and others},
  journal = {arXiv preprint arXiv:2412.00207},
  year    = {2024},
  url     = {https://arxiv.org/abs/2412.00207}
}

@article{hu2025simbench,
  title   = {{SimBench}: Benchmarking the Ability of Large Language Models to Simulate Human Behaviors},
  author  = {Hu, Tiancheng and Baumann, Joachim and Lupo, Lorenzo and Hovy, Dirk and Collier, Nigel and R{\"o}ttger, Paul},
  journal = {arXiv preprint arXiv:2510.17516},
  year    = {2025},
  url     = {https://arxiv.org/abs/2510.17516}
}

@article{argyle2023out,
  title   = {Out of One, Many: Using Language Models to Simulate Human Samples},
  author  = {Argyle, Lisa P. and Busby, Ethan C. and Fulda, Nancy and Gubler, Joshua R. and Rytting, Christopher and Wingate, David},
  journal = {Political Analysis},
  volume  = {31},
  number  = {3},
  pages   = {337--351},
  year    = {2023},
  doi     = {10.1017/pan.2023.2}
}

@article{liu2026humanstudy,
  title   = {{HumanStudy-Bench}: Towards {AI} Agent Design for Participant Simulation},
  author  = {Liu, Xuan and Shang, Haoyang and Liu, Zizhang and Liu, Xinyan and Xiao, Yunze and Tu, Yiwen and Jin, Haojian},
  journal = {arXiv preprint arXiv:2602.00685},
  year    = {2026},
  url     = {https://arxiv.org/abs/2602.00685}
}

@inproceedings{voss2022multimodal,
  title     = {Addressing Data Scarcity in Multimodal User State Recognition
               by Combining Semi-Supervised and Supervised Learning},
  author    = {Vo{\ss}, Hendric and Wersing, Heiko and Kopp, Stefan},
  booktitle = {Companion Publication of the Int. Conf. on Multimodal
               Interaction (ICMI)},
  year      = {2021},
  url       = {https://arxiv.org/abs/2202.03775}
}

@inproceedings{wachowiak2024,
  title     = {Are Large Language Models Aligned with People's Social Intuitions
               for Human-Robot Interactions?},
  author    = {Wachowiak, Lennart and Coles, Andrew and Celiktutan, Oya
               and Canal, Gerard},
  booktitle = {Proc. IEEE/RSJ Int. Conf. Intelligent Robots and Systems (IROS)},
  year      = {2024},
  url       = {https://arxiv.org/abs/2403.05701}
}

@inproceedings{kim2024understanding_llm_hri,
  title     = {Understanding {Large-Language Model} ({LLM})-Powered
               Human-Robot Interaction},
  author    = {Kim, Callie Y. and Lee, Christine P. and Mutlu, Bilge},
  booktitle = {Proc. ACM/IEEE Int. Conf. Human-Robot Interaction (HRI)},
  pages     = {371--380},
  year      = {2024},
  doi       = {10.1145/3610977.3634966},
  url       = {https://dl.acm.org/doi/10.1145/3610977.3634966}
}

@inproceedings{garello2025building,
  title     = {Building Knowledge from Interactions: An {LLM}-Based
               Architecture for Adaptive Tutoring and Social Reasoning},
  author    = {Garello, Luca and Belgiovine, Giulia and Russo, Gabriele
               and Rea, Francesco and Sciutti, Alessandra},
  booktitle = {Proc. IEEE/RSJ Int. Conf. Intelligent Robots and
               Systems (IROS)},
  year      = {2025},
  url       = {https://arxiv.org/abs/2504.01588}
}

@inproceedings{zhao2023chat_environment,
  title     = {Chat with the Environment: Interactive Multimodal
               Perception Using Large Language Models},
  author    = {Zhao, Xufeng and Li, Mengdi and Weber, Cornelius
               and Hafez, Muhammad Burhan and Wermter, Stefan},
  booktitle = {Proc. IEEE/RSJ Int. Conf. Intelligent Robots and
               Systems (IROS)},
  pages     = {3590--3596},
  year      = {2023},
  doi       = {10.1109/IROS55552.2023.10342363},
  url       = {https://ieeexplore.ieee.org/document/10342363/}
}

@inproceedings{kwon2023grounded,
  title     = {Toward Grounded Commonsense Reasoning},
  author    = {Kwon, Minae and Hu, Hengyuan and Myers, Vivek
               and Karamcheti, Siddharth and Dragan, Anca
               and Sadigh, Dorsa},
  booktitle = {Proc. IEEE Int. Conf. Robotics and Automation (ICRA)},
  pages     = {5463--5470},
  year      = {2024},
  url       = {https://arxiv.org/abs/2306.08651}
}

@article{hwang2025tomagent,
  title   = {Infusing Theory of Mind into Socially Intelligent {LLM} Agents},
  author  = {Hwang, EunJeong and others},
  journal = {arXiv preprint arXiv:2509.22887},
  year    = {2025},
  url     = {https://arxiv.org/abs/2509.22887}
}

@article{xu2024sesi,
  title   = {Academically Intelligent {LLMs} are Not Necessarily Socially Intelligent},
  author  = {Xu, Ruoxi and others},
  journal = {arXiv preprint arXiv:2403.06591},
  year    = {2024},
  url     = {https://arxiv.org/abs/2403.06591}
}

@inproceedings{zhou2023sotopia,
  title     = {{SOTOPIA}: Interactive Evaluation for Social Intelligence
               in Language Agents},
  author    = {Zhou, Xuhui and Zhu, Hao and Mathur, Leena and Zhang, Ruohong
               and Yu, Haofei and Qi, Zhengyang and Morency, Louis-Philippe
               and Bisk, Yonatan and Fried, Daniel and Neubig, Graham and others},
  booktitle = {Proc. Int. Conf. Learning Representations (ICLR)},
  year      = {2024},
  url       = {https://arxiv.org/abs/2310.11667}
}

@article{irfan2025between,
  title   = {Between Reality and Delusion: Challenges of Applying Large Language
             Models to Companion Robots for Open-Domain Dialogues with Older Adults},
  author  = {Irfan, Bahar and Kuoppam{\"a}ki, Sanna and Hosseini, Aida
             and Skantze, Gabriel},
  journal = {Autonomous Robots},
  volume  = {49},
  number  = {1},
  pages   = {9},
  year    = {2025},
  publisher = {Springer}
}

@article{bartneck2009godspeed,
  title   = {Measurement Instruments for the Anthropomorphism, Animacy, Likeability,
             Perceived Intelligence, and Perceived Safety of Robots},
  author  = {Bartneck, Christoph and Kuli{\'c}, Dana and Croft, Elizabeth
             and Zoghbi, Susana},
  journal = {International Journal of Social Robotics},
  volume  = {1},
  number  = {1},
  pages   = {71--81},
  year    = {2009},
  doi     = {10.1007/s12369-008-0001-3}
}

@inproceedings{carpinella2017rosas,
  title     = {The Robotic Social Attributes Scale ({RoSAS}): Development and Validation},
  author    = {Carpinella, Colleen M. and Wyman, Alisa B. and Perez, Michael A.
               and Stroessner, Steven J.},
  booktitle = {Proc. ACM/IEEE Int. Conf. Human-Robot Interaction (HRI)},
  pages     = {254--262},
  year      = {2017},
  doi       = {10.1145/2909824.3020208}
}

@article{irfan2024cues,
  title   = {Human-Robot Interaction Conversational User Enjoyment
             Scale ({HRI CUES})},
  author  = {Irfan, Bahar and Miniota, Jura and Thunberg, Sofia
             and Lagerstedt, Erik and Kuoppam{\"a}ki, Sanna
             and Skantze, Gabriel and Pereira, Andr{\'e}},
  journal = {IEEE Transactions on Affective Computing},
  year    = {2025},
  doi     = {10.1109/TAFFC.2025.3590359},
  url     = {https://ieeexplore.ieee.org/document/11084909/}
}

@misc{irfan2024dataset,
  title        = {{HRI CUES} Dataset (Anonymized)},
  author       = {Irfan, Bahar and others},
  year         = {2024},
  howpublished = {Zenodo},
  doi          = {10.5281/zenodo.12588810},
  url          = {https://zenodo.org/records/12588810},
  note         = {CC-BY 4.0}
}

@article{stroessner2019social,
  title     = {The Social Perception of Humanoid and Non-Humanoid Robots:
               Effects of Gendered and Machinelike Features},
  author    = {Stroessner, Steven J. and Benitez, Jonathan},
  journal   = {International Journal of Social Robotics},
  volume    = {11},
  number    = {2},
  pages     = {305--315},
  year      = {2019},
  publisher = {Springer}
}

@article{reeves2020social,
  title     = {Social Robots Are Like Real People: First Impressions, Attributes,
               and Stereotyping of Social Robots},
  author    = {Reeves, Byron and Hancock, Jeff and Liu, Xun},
  journal   = {Technology, Mind, and Behavior},
  volume    = {1},
  number    = {1},
  pages     = {76},
  year      = {2020},
  publisher = {American Psychological Association}
}

@article{matuszek2026reporting,
  author    = {Matuszek, Cynthia and Williams, Tom and DePalma, Nicholas
               and Mead, Ross and Wen, Ruchen and Schneiders, Eike
               and Kennington, Casey and Bezabih, Alemitu},
  title     = {Reporting Guidelines for Large Language Models in
               Human--Robot Interaction},
  year      = {2026},
  publisher = {Association for Computing Machinery},
  address   = {New York, NY, USA},
  volume    = {15},
  number    = {2},
  url       = {https://doi.org/10.1145/3777552},
  doi       = {10.1145/3777552},
  journal   = {J. Hum.-Robot Interact.},
  month     = jan,
  articleno = {36},
  numpages  = {24}
}

@article{kulic2007affective,
  title   = {Affective State Estimation for Human-Robot Interaction},
  author  = {Kuli{\'c}, Dana and Croft, Elizabeth A.},
  journal = {IEEE Transactions on Robotics},
  volume  = {23},
  number  = {5},
  pages   = {991--1000},
  year    = {2007}
}

@article{song2024vlmsocialnav,
  title   = {{VLM-Social-Nav}: Socially Aware Robot Navigation Through Scoring
             Using Vision-Language Models},
  author  = {Song, Dongkyu and Liang, Jungseok and Payandeh, Amirhossein
             and Raj, Aman Harishkumar and Xiao, Xuesu and Manocha, Dinesh},
  journal = {IEEE Robotics and Automation Letters},
  year    = {2024}
}

@article{amirova2021nao,
  title   = {10 Years of Human-{NAO} Interaction Research: A Scoping Review},
  author  = {Amirova, Aida and Rakhymbayeva, Nazerke and Yadollahi, Elmira
             and Sandygulova, Anara and Johal, Wafa},
  journal = {Frontiers in Robotics and AI},
  volume  = {8},
  pages   = {744526},
  year    = {2021}
}

@inproceedings{radford2023whisper,
  title     = {Robust Speech Recognition via Large-Scale Weak Supervision},
  author    = {Radford, Alec and Kim, Jong Wook and Xu, Tao and Brockman, Greg
               and McLeavey, Christine and Sutskever, Ilya},
  booktitle = {Proceedings of the 40th International Conference on Machine
               Learning (ICML)},
  pages     = {28492--28518},
  year      = {2023}
}

@article{ullman2023tom,
  title   = {Large Language Models Fail on Trivial Alterations to Theory-of-Mind Tasks},
  author  = {Ullman, Tomer},
  journal = {arXiv preprint arXiv:2302.08399},
  year    = {2023}
}

@inproceedings{sharma2024sycophancy,
  title     = {Towards Understanding Sycophancy in Language Models},
  author    = {Sharma, Mrinank and Tong, Meg and Korbak, Tomasz and Duvenaud, David
               and Askell, Amanda and Bowman, Samuel and others},
  booktitle = {Proceedings of the International Conference on Learning
               Representations (ICLR)},
  year      = {2024}
}

@inproceedings{laban2025people,
  title        = {What People Share with a Robot When Feeling Lonely and Stressed
                  and How It Helps over Time},
  author       = {Laban, Guy and Chiang, Sophie and Gunes, Hatice},
  booktitle    = {2025 34th IEEE International Conference on Robot and Human
                  Interactive Communication (RO-MAN)},
  pages        = {1930--1935},
  year         = {2025},
  organization = {IEEE}
}

@article{amrhein2019scientists,
  title     = {Scientists Rise Up Against Statistical Significance},
  author    = {Amrhein, Valentin and Greenland, Sander and McShane, Blake},
  journal   = {Nature},
  volume    = {567},
  number    = {7748},
  pages     = {305--307},
  year      = {2019},
  publisher = {Nature Publishing Group UK London}
}

@inproceedings{hutto2014vader,
  title     = {{VADER}: A Parsimonious Rule-Based Model for Sentiment Analysis
               of Social Media Text},
  author    = {Hutto, Clayton and Gilbert, Eric},
  booktitle = {Proceedings of the International AAAI Conference on Web
               and Social Media},
  volume    = {8},
  number    = {1},
  pages     = {216--225},
  year      = {2014}
}

@article{singh2024regulating,
  title     = {Regulating Modality Utilization Within Multimodal Fusion Networks},
  author    = {Singh, Saurav and Saber, Eli and Markopoulos, Panos P.
               and Heard, Jamison},
  journal   = {Sensors},
  volume    = {24},
  number    = {18},
  pages     = {6054},
  year      = {2024},
  publisher = {MDPI}
}

\end{document}